\def\year{2022}\relax
\documentclass[letterpaper]{article} 
\usepackage{aaai22}  
\usepackage{times}  
\usepackage{helvet}  
\usepackage{courier}  
\usepackage[hyphens]{url}  
\usepackage{graphicx} 
\usepackage{float}

\usepackage{amsmath,amssymb}
\usepackage{natbib}  
\usepackage{caption} 
\DeclareCaptionStyle{ruled}{labelfont=normalfont,labelsep=colon,strut=off} 
\usepackage[ruled,vlined]{algorithm2e}
\usepackage{newfloat}
\usepackage{subcaption}
\usepackage{listings}
\usepackage{booktabs}
\usepackage{multirow}
\floatstyle{ruled}
\newfloat{listing}{tb}{lst}{}
\floatname{listing}{Listing}
\title{Unsupervised Driving Behavior Analysis using Representation Learning and Exploiting Group-based Training}
\author {
    Soma Bandyopadhyay,\textsuperscript{\rm 1}
    Anish Datta, \textsuperscript{\rm 1}
    Shruti Sachan, \textsuperscript{\rm 1}
    Arpan Pal \textsuperscript{\rm 1}
}
\affiliations {
    \textsuperscript{\rm 1} TCS Research, TATA Consultancy Services ,Kolkata, India \\
    soma.bandyopadhyay@tcs.com, anish.datta@tcs.com, sachan.shruti@tcs.com, arpan.pal@tcs.com
}

\usepackage{bibentry}

\begin{document}

\maketitle

\begin{abstract}
Driving behavior monitoring plays a crucial role in managing road safety and decreasing the risk of traffic accidents. Driving behavior is affected by multiple factors like vehicle characteristics, types of roads, traffic, but, most importantly, the pattern of driving of individuals. Current work performs a robust driving pattern analysis by capturing variations in driving patterns. It forms consistent groups by learning compressed representation of time series (Auto Encoded Compact Sequence) using a multi-layer seq-2-seq autoencoder and exploiting hierarchical clustering along with recommending the choice of best distance measure. Consistent groups aid in identifying variations in driving patterns of individuals captured in the dataset. These groups are generated for both train and hidden test data. The consistent groups formed using train data, are exploited for training multiple instances of the classifier. Obtained choice of best distance measure is used to select the best train-test pair of consistent groups. We have experimented on the publicly available UAH-DriveSet dataset considering the signals captured from IMU sensors (accelerometer and gyroscope) for classifying driving behavior. We observe proposed method, significantly outperforms the benchmark performance.
\end{abstract}

\section{Introduction}

Driving behavior analysis is a growing field of research due to increasing safety concerns in roads and the advent of semi-autonomous vehicles. According to World Health Organization (WHO), traffic accidents are responsible for approximately 1.3 million deaths in a year and incurs almost 3\% cost of most countries GDP \cite{who}. Majority of these accidents occurs due to inattentive/distracted and aggressive/rash driving behavior. Therefore, identifying category of driving patterns correctly helps to avoid unwanted accidents and can become a key player to restore road safety. 

One of the main challenges for behavior analysis, is the existence of diversity in individuals, depending on individual’s inherent characteristics and, also based on demographic factors such as age, gender etc. Hence, building a model which would be addressing this diversity in the training data, as well as, for inferencing is an important requirement. We address this requirement, by proposing a method for classification of driving behavior using representation learning \cite{bengio2013} and group-based training. A consistent group formation technique is formulated using Hierarchical Clustering (HC) using Auto-encoded Compact Sequence, a prior work on HC \cite{bandyopadhyay2021,luczak2016,friedman2017}, to identify the variations of driving patterns in diverse individuals. This method is performed iteratively, until we observe a consistent set of groups which cannot be further divided into subgroups. Separate instances of a classification model are trained for each consistent group formed in the training set. In the hidden test data, similar group formation is performed, and the groups are mapped for inferencing to the train model exploiting the best distance measure.    

There exist few prior works on clustering-based ensemble of classifiers \cite{rahman2013,jurek2014,chakraborty2017} to learn the patterns of each cluster separately using multiple models. But these techniques are not robust, as it is very difficult to inherently assume the number of groups to be formed for a specific dataset. Our method deals with this challenge by automatically finding the number of consistent groups in a dataset without any prior assumptions. Also, none of these prior works have applied representation learning to retain the informative part of the data, as well as, achieve the group distribution using learned representation with a choice of best clustering and distance measure.

Our prime contributions are as follows:\\
1.	\textbf{Consistent group formation using representation learning and choice of best distance measure (CGRL)}: Iterative grouping of data is performed until no further separation is possible i.e., they become consistent. Hierarchical Clustering using Auto-encoded Compact Sequence have been used for grouping the data to capture the subtle variations of driving pattens and divide into different clusters. A compact representation of the time-series is learnt using multi-layer seq-2-seq auto-encoder and subsequently, hierarchical clustering is performed on the learned representation using the selected choice of best distance measure among Chebyshev, Manhattan and Mahalanobis distance. Consistent groups are formed for both the train and test data. \\
2.	\textbf{Learning using consistent groups}: Separate instances of a base classification model are trained for each of the consistent train groups.  \\
3.	\textbf{Mapping best pair of train - test group combinations}: The test groups are inferred using the classification model using two different techniques. One uses the consistent group representative (CR) distance between the train and test groups, and another exploits the average instance-wise distance between the groups. The distances are computed in the representation space using the recommended best distance measure.  \\ 
4.	\textbf{Extensive analysis on publicly available driving behavior analysis dataset}: We evaluated our method on publicly available UAH-Driveset data \cite{romera2016}, which consists of three driving behaviors- Normal, Aggressive and Drowsy. Data from six drivers with varying age-groups, gender and vehicle type have been collected using a smartphone. IMU signals i.e acceleration and angular velocity have been considered for experimentation. Proposed method outperforms benchmark classification models like MLSTM-FCN and Stacked-LSTM (trained using a single model) by 5\% and 2\% respectively, with a significant improvement in aggressive driving behavior prediction. \\

\section{Related Works}

In recent years, there have been significant research related to driving behavior analysis. Primarily, data collected from sensors inbuilt in vehicles or via smartphones are used for performing such analysis. Broadly, time-series data from vehicles are collected from two types of sensors: (1) Location sensors like GPS etc. and (2) Inertial Measurement sensors (IMU) sensors like accelerometer, gyroscope etc.  				

\cite{wang2018} used GPS-trajectory data to develop a framework for driving behavior analysis. Firstly, sequence of transition graphs with driving states are computed using GPS data. Subsequently, an autoencoder-based model is used to learn the latent representations of driving behavior from the transition graphs. \cite{khodairy2021} proposed an optimized Stacked-LSTM model on signals collected from a smartphone for driving behavior analysis to classify normal. aggressive and drowsy behavior. \cite{zhang2019} proposed a deep learning-based framework that automatically learns rich feature representations of driving behaviors and captures salient features from high-dimensional sensor data by fusing convolutional neural networks and recurrent neural networks along with an attention mechanism for in-vehicle CAN-BUS sensor data. 

\cite{guo2018} developed a hybrid unsupervised deep learning model (AESOM) combining autoencoder to extract latent features and self-organized maps for classifying driving behavior applied to data collected from GPS sensors. \cite{fugiglando2018} focused on signals from CAN Bus Data that are directly or indirectly related to interaction between the driver and the vehicle, and, grouped driver's behavior using K-means clustering and cross validation. The analysis showed that specific combinations of a signal and feature provide the most discriminating signatures between driving behaviors. \cite{moosavi2021} presented a deep-neural-network architecture, termed D-CRNN, for capturing representations for driving style, using CNN and RNN. CNN captures the semantic patterns of driving behavior and temporal dependencies between them are learnt using RNN.

\section{Proposed Method}

\subsection{Consistent group formation using representation learning and choice of best distance measure}

\subsubsection{Hierarchical Clustering using Auto-Encoded Compact Sequence (HC-AECS):}

\begin{figure}[htbp]
\centering
\includegraphics[width=0.8\columnwidth]{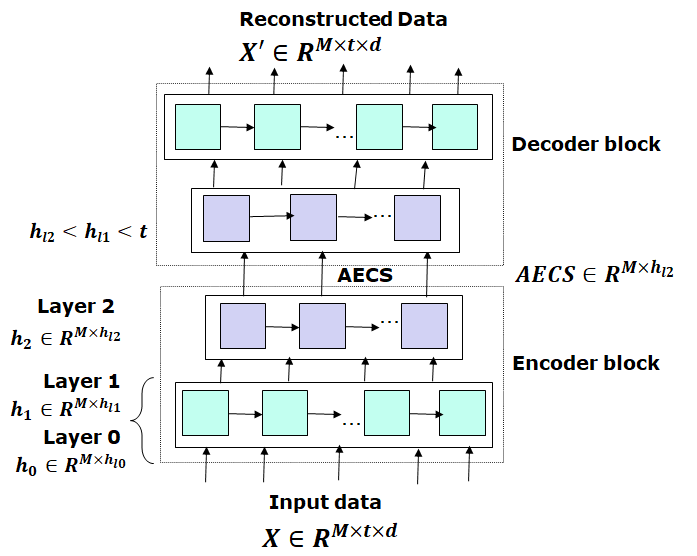}
\caption{Schematic block diagram of multi-layer seq-2-seq autoencoder for computing AECS}
\label{aecs_archi}
\end{figure}

It uses representation learning for robust and efficient time-series clustering to obtain a choice of best clustering and associated distance measure. In this method, first a compact representation (AECS) of time-series is learned using a seq-2-seq LSTM undercomplete auto-encoder \cite{hochreiter1997}. Figure \ref{aecs_archi} depicts the auto-encoder architecture where M, t and d are the number of instances, time-steps and dimensions of input data respectively. Here $h_{l1}$ and $h_{l1}$ indicates the number of nodes in the hidden layers $l1$ and $l2$ respectively.

\begin{figure*}[htbp]
\centering
\includegraphics[width=0.8\textwidth]{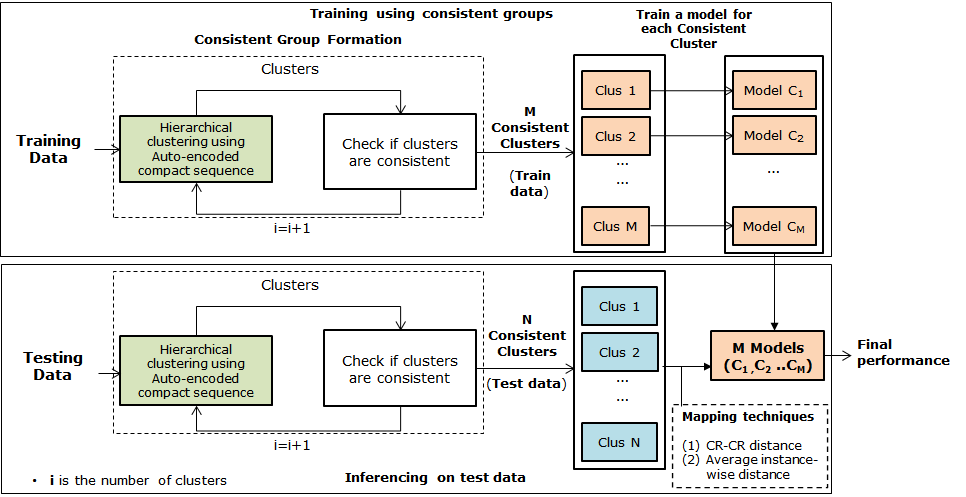}
\caption{Functional block diagram of proposed method}
\label{block}
\end{figure*}

After AECS is learnt, agglomerative hierarchical clustering is performed on this compact representation using a method to find the best choice of distance measure and associated best clustering. Three distance measures have been used to evaluate the distance between two time-series for hierarchical clustering. The distance measures used are Chebyshev Distance, Manhattan Distance \cite{wang2013} and Mahalanobis Distance \cite{de2000}. 

Suppose $T_i$= \{$t_i^1,t_i^2,..t_i^k,..,t_i^n$\} and $T_j$= \{$t_j^1,t_j^2,..t_j^k,..,t_j^n$\} be two time-series of length n. The distance measures can be defined as: \\
(1) Chebyshev distance:     $max_k (|t_i^k-t_j^k |)$    \\
(2) Manhattan distance:     $\sum_{k=1}^n |t_i^k-t_j^k |$ \\                                    (3) Mahalanobis distance: $\sqrt{(T_i-T_j)^T.C^{-1}.(T_i-T_j)^T}$  \\
where $C$ is the co-variance matrix between $T_i$ and $T_j$. 

The clustering obtained by the 3 different distance measures are compared using Modified Hubert Statistic ($\rho$) \cite{hubert1985}, an internal clustering measure, for identification of the best distance measure. $\rho$ measures the separation between the clusters, exploiting pairwise instances distance weighted by the distance between their centroids. The distance measure using which the clustering produces maximum value of Modified Hubert Statistic is selected as best distance measure.

\subsubsection{Consistent Group Formation (CGF):}

Here, we devise a method to discover the inherent groups in a dataset without using any annotations or prior knowledge. 
We have formed a concept of consistent grouping where the generated groups cannot be broken further into subgroups (comprising of significant number of instances). The groups are formed using HC-AECS, exploiting agglomerative hierarchical clustering on the learnt compact representation by using multi-layer undercomplete auto-encoder. It iteratively performs clustering on the learned representation, increasing the number of clusters to be formed by one in each successive iteration (starting from 2 clusters). At any iteration, if the maximum number of instances / data points are retained in the same groups as in previous iteration, the process is stopped, and the previous set of groups are termed as consistent. We define a threshold $\tau$ as the stopping criterion, to track if the new cluster formed at an iteration has lower than $\tau$ fraction of the instances of the dataset. It infers the new group formed is very small in size, and the set of previous groups are returned. We form the consistent groups for both training and testing data. Algorithm \ref{cgf} describes the method for consistent group formation.

\begin{algorithm}[h]
\SetKwInOut{Input}{Input}
\SetKwInOut{Output}{Output}
\Input{$D$: Time-series, \\
$\tau$: Threshold (stopping criterion)}
\Output{$CG$: Consistent groups}
\SetAlgoLined
\caption{Consistent Group Formation (CGF)}
\label{cgf}
\SetKwProg{generate}{Function \emph{CGF}}{}{end}
\generate{($D$, $\tau$)}{
	 $k$ $\gets$ 2 ; $\triangleright$ Initializing number of clusters \\
	 $N$ $\gets$ Number of time-series in $D$\;
	 $Stopping\_cond$ $\gets$ False \; 
	 $C_{k-1} \gets $ [] \;
	 \While{$Stopping\_cond = False$}{
	 $C_k , d_k \gets HC-AECS(D, k)$ \;     
	 \If{Difference $(C_k  , C_{k-1} ) < \tau * N$}{
	 $Stopping\_cond \gets True$ \;
	 $CG \gets C_{k-1}$ ; $\triangleright$ Consistent clusters \\
	 \textbf{return} $CG$	 \;	 
	 }
	 $C_{k-1} \gets C_k$ \;
	 $k \gets k + 1$ ; $\triangleright$ Increment no. of clusters by 1
	 }
}
\end{algorithm}

\subsection{Learning using consistent clusters}

After the consistent clusters are formed for the training data, separate instances of a classification model are trained on each of the consistent clusters formed. As each consistent group have different dominant demographic characteristics of the drivers, training a model on each of the groups separately is expected to improve the efficacy of the learner. 
Suppose $K$ consistent groups are formed for the training set and $C$ be the baseline classifier used. Here, $i^{th}$ instance of classifier $C$ is trained using the data in consistent group i, ($X_i$) and its corresponding labels $y_i$ as shown in equation 4.

\begin{equation}
C_i \; = \; Train (C,(X_i,y_i)),\; where \; 1 \leq i \leq K         
\end{equation}

\subsection{Inferencing using the consistent test groups exploiting the best distance measure}

We perform consistent grouping on the test data and subsequently, consistent test groups are mapped w.r.t best training model without using any annotation. We propose two different methods for mapping the test groups to the optimal trained classifier for inferencing. \\
1. Using Consistent group representative distance (CR-CR distance) \\
2. Using average instance-wise distance (Avg distance)

\begin{figure}[htbp]
\centering
\includegraphics[width=0.7\columnwidth]{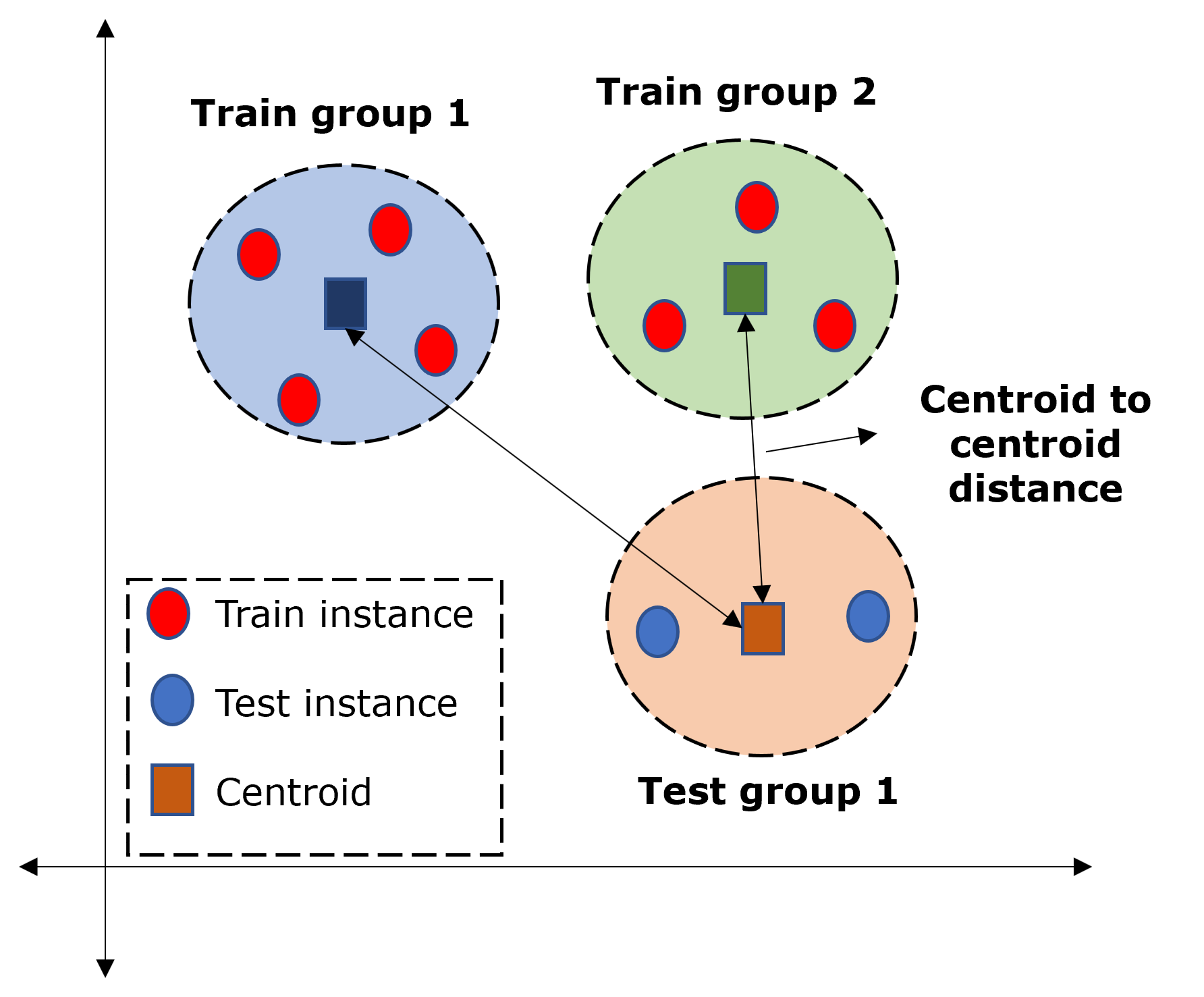}
\caption{Depiction of consistent group representative (CR) distance between a test group and 2 train groups}
\label{cen_to_cen}
\end{figure}

(1) \textbf{Using CR to CR distance:} We compute the consistent group representatives (CR) for each of the groups in the latent representation space i.e. using AECS. The CRs are computed as the mean of the elements present in the group as depicted in Figure \ref{cen_to_cen}. Similarly, we also find the CRs of the groups formed in train set using AECS. We map each test group, to the model trained using the group, whose distance of the group representative i.e. CR, is lowest to the CR of the test group computed using the best distance measure.  
Let $CR_i$ be the representative of training group $i$ , $CR_j$ the representative for test group $j$ in the representation space and $d$ be the best distance measure.  Then model trained with training group $k$ i.e. $C_k$ is selected for test group $j$ where,

\begin{equation}
k=ArgMin_i \; d(CR_i  ,CR_j ), \; 1 \leq i \leq K 
\label{cen_dis_eq}   
\end{equation}
                 
(2) \textbf{Using average instance-wise distance:} In this method, we compute the distance between each instance in a test group to each instance to a train group in the representation space and find their average. The distances are computed using the best distance measure. For each test group, the model is chosen trained using the group which has minimum average instance-wise distance from the test group.

\begin{figure}[!h]
\centering
\includegraphics[width=0.75\columnwidth]{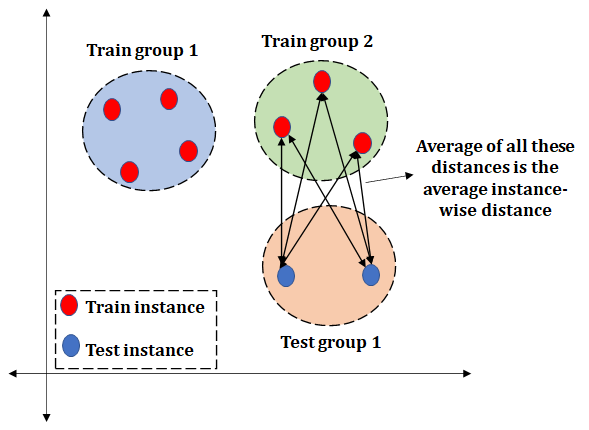}
\caption{Depiction of average instance-wise distance between a test group and train group}
\label{avg_dis}
\end{figure}

Suppose $i$ = \{$a_1,a_2…,a_n$\} be a training group and $j$ = \{$b_1,b_2…,b_m$\} be a test group. The average instance-wise distance between the two groups as shown in Figure \ref{avg_dis} can be defined as:
\begin{equation}
Avg\_dist(i,j)=\frac{1}{nm} \sum_{i=1,j=1}^{i \leq n,j \leq m} d(a_i,b_j) ,  
\end{equation}
where $d$ is the best distance measure chosen. 

For any test group $j$, the model is used for inferencing whose training group ($k$) has minimum $Avg\_dist$ w.r.t $j$.

\begin{equation}
k=ArgMin_i \;  Avg\_dist(i,j), \; 1 \leq i \leq K 
\label{avg_dis_eq}                
\end{equation}

\begin{algorithm}[h]
\SetKwInOut{Input}{Input}
\SetKwInOut{Output}{Output}
\Input{$\{X_{train},y_{train}\}$: Training set, \\
$X_{test}$: Test set, \\
$C$: Base classifier}
\Output{$y_{test}$: Annotations of test data}
\SetAlgoLined
\caption{Learning and Inferencing using Consistent Groups}
\label{proposed}
\SetKwProg{generate}{Function \emph{Training\_using\_CG}}{}{end}
\generate{($X_{train}$, $y_{train}$)}{
	 $\triangleright$ Find consistent groups in train set \\
	 $CG_{train} \gets CGF (X_{train})$ ; $\triangleright$ From Algorithm 1 \\
	 $K_{train} \gets$ Number of groups in $CG_{train}$ \;
	 $\triangleright$ Train a model for each consistent cluster $i$ \\
	 \ForAll{$i \in 1,2,.. ,K_{train}$}{
	 $ind_i \gets$ Instances where $CG_{train} = i$ \;
	 $C_i \gets Train(C(X_{train}[ind_i], y_{train}[ind_i]))$\;
	 }
	 \textbf{return} trained classifiers $\{C_1, C_2, … , C_{K_{train}}\}$\;
}
\SetKwProg{generate}{Function \emph{Inferencing\_using\_CG}}{}{end}
\generate{($X_{test}$)}{
	 $\triangleright$ Find consistent groups in test set \\
	 $CG_{test} \gets CGF (X_{test})$ ; $\triangleright$ From Algorithm 1 \\
	 $K_{test} \gets$ Number of groups in $CG_{test}$ \;
	 $\triangleright$ Perform inference on each consistent cluster $i$ \\
	 \ForAll{$i \in 1,2,.. ,K_{test}$}{
	 $ind_i \gets$ Instances where $CG_{test} = i$ \;
	 $b \gets$ Mapping test group $i$ to best train group using equations \ref{cen_dis_eq} or \ref{avg_dis_eq} \;
	 $\triangleright$ Infer test group using $b^{th}$ classifier $C_b$ \\
	 $y_{test}[ind_i] \gets Infer(C_b (X_{test}[ind_i]))$ \; 
	 }
	 \textbf{return} $y_{test}$\;
}
\end{algorithm}

The functional block diagram of proposed approach is described in Figure \ref{block}. Firstly, $M$ consistent clusters (Clus $1$,...,Clus $M$) are formed on the training data using the CGF method as described in Algorithm 1. Then, separate instances of classification models ($C_1$, ...,$C_M$) are trained for each $M$ consistent clusters ($Training\_using\_CG$ function of Algorithm 2). For inferencing, $N$ consistent clusters (Clus $1$,...,Clus $N$) are formed on the test data and the best train-test group combinations are selected for each of $N$ clusters using any of the two proposed mapping techniques, CR-CR distance and Avg dist ($Inferencing\_using\_CG$ function of Algorithm 2).

\section{Experimental Analysis}

\subsection{Dataset Description}

We evaluate our methodology on a publicly available driving behavior analysis dataset - UAH Driveset data \cite{romera2016}. Time-series recordings of more than 500 minutes are captured by driving monitoring app DriveSafe \cite{bergasa2014} using smartphone-embedded sensors (e.g., inertial measurement, GPS, and cameras). Independent driving sessions were conducted on 6 different drivers with different ages and vehicles. The description of the demographic information of the drivers and their vehicle’s fuel type is depicted in Table \ref{driver}. Every driver performed a series of different behaviors: normal, aggressive and, drowsy and driving on two types of roads (motorway and secondary). The dataset provides a variety of sensor signals, however, in this work, we are interested in raw inertial measurements signals. This is because, the raw inertial signals like acceleration etc. are most representative of the driver’s behavior patterns than other signals like GPS etc.   

\begin{table}
\centering
\begin{tabular}{c c c c}
\hline
\textbf{Driver}	& \textbf{Age range} & \textbf{Gender} &	\textbf{Fuel type} \\ \hline
D1	& 40-50 &	Male & 	Diesel \\ 
D2	& 20-30	& Male	& Diesel \\ 
D3	& 20-30	& Male	& Diesel \\ 
D4	& 30-40	& Female	& Gasoline\\
D5	& 30-40	& Male	& Gasoline\\
D6	& 40-50	& Male	& Electric\\
\hline
\end{tabular}
\caption{Demographic and vehicle information of drivers in UAH Drive-set data}
\label{driver}
\end{table}

\subsection{Data Preprocessing}

The UAH-DriveSet dataset consists of signals collected from smartphone embedded sensors: inertial measurements, GPS, and cameras that are sampled at different frequencies (10 Hz, 1 Hz, 10 Hz, respectively). As mentioned earlier, we only use the inertial measurements i.e. acceleration, roll, pitch and yaw for motorway road. For experimentation, Kalman filtered acceleration signal is used to reduce the impact of external noise from the sensors already provided in the given data. 

Driving sessions were performed with a specific driving behavior (normal, aggressive, or drowsy) for the six drivers. Each session is segmented into fixed sliding windows (segments) of 64 timesteps with a 50\% overlap and then labeled according to the associated driving behavior. Subsequently, all the constructed time-series windows are combined in a single dataset. Finally, we use random splitting technique to split the combined dataset into training and testing sets with ratios of 80\% and 20\%, respectively. The split is performed in a stratified way, based on the combination of driver information, and driving behavior. The description of the pre-processed UAH-Driveset data is provided in Table \ref{prep}. 

\begin{table}[htp]
\centering
\begin{tabular}{c c}
\hline
\textbf{Parameter}	& \textbf{Value} \\ \hline
Train set size	& 4184 \\
Test set size	& 1046 \\
Timesteps	& 64\\
Classes	& 3 (Normal, Drowsy, Aggressive)\\
Features	& 6 (AccX, AccY, AccZ, Roll, Pitch, Yaw)\\
\hline
\end{tabular}
\caption{Description of pre-processed UAH-Driveset Data}
\label{prep}
\end{table}


\subsection{Representation learning with consistent groups formation}

In this section, we elaborate the mechanism for forming consistent groups on the UAH-Driveset data along with learned representation. 

\subsubsection{Representation Learning:}

Suppose $X_{train}$ be the training set, where $X_{train} \in \mathcal{R}^ {4184 \times 64 \times 6}$. In the seq-2-seq autoencoder, for learning the latent representation, a 2-layer LSTM network is used, where number of units of the two hidden layers are 16 and 12 respectively. Here, for $X_{train}$, a compact representation $AECS_{train} \in \mathcal{R}^ {4184 \times 12}$ is learnt. Performance of afore-mentioned model configuration has been analyzed, tested and validated across diverse uni-variate and multi-variate time-series \cite{bandyopadhyay2021}.

\begin{figure*}[htp]
\label{tsne}
\centering
\begin{subfigure}{.35\textwidth}  
\centering
  \includegraphics[width=\textwidth]{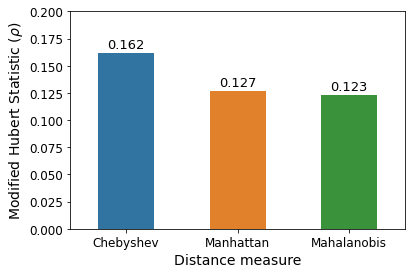}
  \caption*{Figure 5(a): Modified Hubert Statistic ($\rho$) value for clustering on train data using the 3 distance measures}
  \captionlistentry{}
  \label{hubert}
\end{subfigure}
\begin{subfigure}{.60\textwidth}
\centering
  \includegraphics[width=0.45\textwidth]{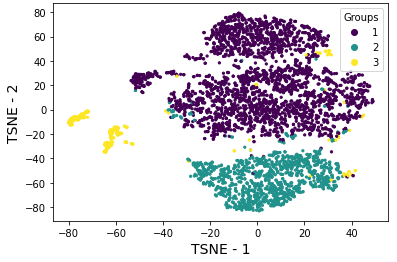}
  \includegraphics[width=0.45\textwidth]{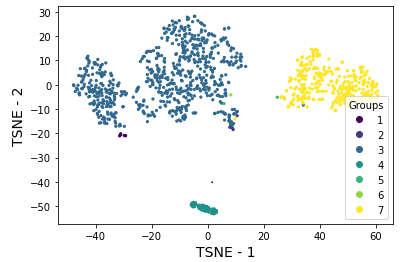}
  \caption*{Figure 5(b):Visualization of consistent train (left) and test groups (right) of UAH-Driveset data using t-SNE in representation space}
  \captionlistentry{}
  \label{train}
\end{subfigure}%


\end{figure*}

\subsubsection{Consistent group formation:} 

Next, we perform Hierarchical clustering on the AECS using 3 distance measures Chebyshev (CH), Manhattan (MA) and Mahalanobis (ML) distance to form consistent groups. We observe Chebyshev distance forms the best clustering, as it produces maximum value of Modified hubert statistic ($\rho$) as depicted in Figure \ref{hubert}. Three consistent groups are formed for the training data, which includes two majority groups and another small outlier group. Similarly, using the proposed method for the test data, a total of seven consistent groups are formed.
%

\subsection{Training on the consistent groups}

On formation of the consistent groups for the training set, we train each of them individually using a benchmark classification model. Raw time-series data is used for training without the need of any hand-crafted features. We have considered two benchmark deep learning models – (1) MLSTM-FCN \cite{karim2019} and (2) Stacked-LSTM model \cite{khodairy2021} for separate experimentations. MLSTM-FCN is an established model for multi-variate time-series classification. We have used for analysis the Stacked-LSTM model, as it is one of the prior works on driving behavior classification, where the authors have experimented with the UAH-Driveset data. Default parameters are used for the models to depict the generalizability of our approach to work on any type of classifier. 

\subsection{Inferencing on the test set}

For each test group, the best train model is selected based on the closeness of the consistent test and train groups formed. Closeness between the groups is measured using the two proposed methods: (1) Consistent group representative (CR) distance, and (2) Average instance-wise distance between test and train groups. The distance used here is the best distance measure obtained while forming the consistent train groups, which is Chebyshev (CH) distance in this case.

Here, the 7 test groups formed are mapped to the closest among the 3 train groups. Subsequently, the instances present in that test group are inferred using the model trained on that train group. Table \ref{mapping} depicts the mapping of the 7 test groups to the optimal training model (denoted by 1,2 and 3). Mapping method indicates the technique used for finding the closeness of the test groups with the training groups (CR-CR: using consistent group representative distance method; Avg: using average instance-wise distance).

\begin{table}[htp]
\centering
\begin{tabular}{|c|c|c|c|c|c|c|c|}
\hline
\textbf{Mapping}	& \multicolumn{7}{|c|}{\textbf{Test Group}} \\ \cline{2-8}
\textbf{Method} & \textbf{1} &	\textbf{2} &	\textbf{3} &	\textbf{4} &	\textbf{5} &	\textbf{6} &	\textbf{7} \\ \hline
CR-CR	& 3	& 3	& 1	& 3	& 1	& 1	& 2 \\
Avg	& 1	& 1	& 1	& 3	& 1	& 1	& 2 \\
\hline
\end{tabular}
\caption{Test group mapping to the optimal training models using two techniques (CR-CR and Avg)}
\label{mapping}
\end{table}

Interestingly, we observe the largest consistent test group (group 3) is mapped to the model trained using the largest training group (group 1) for both the methods. Similarly, the second largest test group (group 7) is mapped to the second largest train group (group 2).

\subsection{Results}

We compare proposed model with the benchmark model trained on the complete training set. We compute the performance of our method using both techniques for mapping the test groups with the optimal training model. Performance metrics like accuracy and F1-score are used to compare our method with benchmark results for classification of normal, drowsy and aggressive behavior. Table \ref{results} depicts the detailed performance of proposed method and comparison with the benchmark model. We observe for both base classifier models Stacked-LSTM and MLSTM-FCN, proposed method outperforms the single benchmark model. A notable improvement in accuracy (5\%) and F1-score (4\%) is observed when MLSTM-FCN is used as the base classifier.

\begin{table}[htp]
\centering
\begin{tabular}{|c|c|c|c|c|}
\hline
\multirow{2}{*}{\textbf{Classifier}}	& \multirow{2}{*}{\textbf{Method}}	& \textbf{Mapping}	& \multirow{2}{*}{\textbf{Acc}}	& \textbf{F1-}\\ 
& & \textbf{Method} & & \textbf{Score} \\ \hline
Stacked LSTM & Benchmark & \multirow{2}{*}{-} & \multirow{2}{*}{0.78} & \multirow{2}{*}{0.78} \\
(Khodiary & -Original &  & & \\ \cline{2-5}
and Abosamra & Proposed & CR-CR & \textbf{0.80} & \textbf{0.80} \\  \cline{3-5}
2021)& method & Avg & \textbf{0.80} & \textbf{0.80} \\ \hline
MLSTM & Benchmark & \multirow{2}{*}{-} & \multirow{2}{*}{0.82} & \multirow{2}{*}{0.82} \\
-FCN & -Original &  & & \\ \cline{2-5}
(Karim et al. & Proposed & CR-CR & \textbf{0.87} & \textbf{0.86} \\ \cline{3-5}
2019) & method & Avg & \textbf{0.87} & \textbf{0.86} \\ \hline
\end{tabular}
\caption{Performance of proposed method and comparison with benchmark model for both base classifiers (Stacked-LSTM and MLSTM-FCN)}
\label{results}
\end{table}

\begin{figure}[ht]
\centering
\begin{subfigure}{.5\columnwidth}
\centering
  \includegraphics[width=\textwidth]{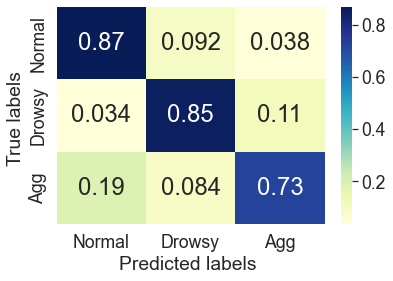}
  \caption{Single model}
  \label{single}
\end{subfigure}%
\begin{subfigure}{.5\columnwidth}  
\centering
  \includegraphics[width=\textwidth]{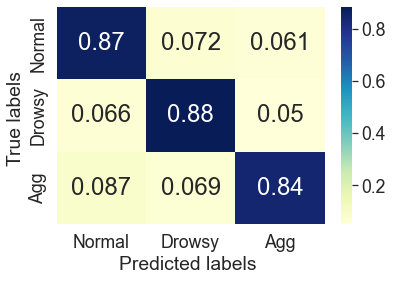}
  \caption{Proposed method}
  \label{prop}
\end{subfigure}
\caption{Confusion matrices of predictions (in acc) on test data using benchmark method and proposed method}
\label{CF}
\end{figure}

Figure \ref{CF} shows the confusion matrices on test data for single model (benchmark) and proposed method (mapping method: avg distance) using MLSTM-FCN as base classifier. We can observe approximately 11\% improvement in prediction of aggressive behavior which is an important aspect for road safety. 

We further observe, both the methods for mapping the test groups to the train models produces the same results and are equally robust.   
We have used t-SNE \cite{van2008} for visualization of the consistent groups formed on the train and test data in latent representation space. Figure \ref{train} depicts the plots showing the separation between the consistent groups for the training and testing data.  

\begin{figure*}[htbp]
\centering
\includegraphics[width=\textwidth]{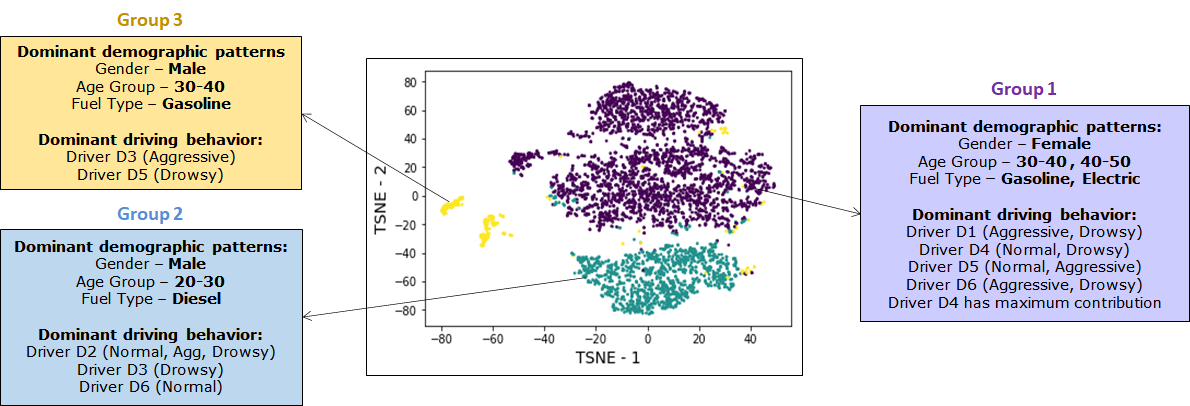}
\caption{Visualization of the 3 consistent training groups and the dominant demographic and driving behavior patterns that characterizes the groups}
\label{dom}
\end{figure*}

\subsection{Analysis of the consistent training groups with driver’s driving behavior}

In this section, we perform analysis on the consistent groups formed in training set using demographic and driving behavior information of the drivers provided in the UAH Driveset data described in Table 1 and 3. 

\subsubsection{Analysis on smallest training group:}

On mapping the instances in the smallest group (group 3 consisting of 253 instances) of the training data, we observe some notable patterns. In Table \ref{outlier}, we depict the window-wise composition of the group 3 w.r.t each of the 6 drivers and their types of driving behavior. We observe, drivers D3 and D5 mainly contribute to this outlier group with 28.86\% and 55.34\% respectively. Furthermore, all instances of driver D3 present in this group are of aggressive class, while majority of the instances (124 of 140) of D5 are drowsy. 

\begin{table}[htp]
\centering
\begin{tabular}{c| c c c |c}
\hline
\multirow{2}{*}{\textbf{Driver}}	& \multicolumn{3}{|c|}{\textbf{Driving Behavior (Group 3)}} & \multirow{2}{*}{\textbf{Total}}\\ \cline{2-4}
& Normal	& Agg &	Drowsy & \\ \hline
D1	& 8	& 5	& 3	& 16 \\
D2	& 2	& 8	& 1	& 11 \\
\textbf{D3}	& 0	& 73	& 0	& \textbf{73} \\
D4	& 1	& 8	& 1	& 10 \\
\textbf{D5}	& 5	& 11	& 124	& \textbf{140} \\
D6	& 0	& 2	& 1	& 3 \\
\hline
\end{tabular}
\caption{Composition of outlier training group (group 3) by driver information and their driving behaviors}
\label{outlier}
\end{table}

During training, we have also experimented by discarding this group to see if the performance varied. We have observed a significant dip in performance (of approximately 4\%), indicating though this group is a kind of outlier group, being the smallest group, however this is informative and contributes immensely to the training process \cite{bandyopadhyay2016}.

\subsubsection{Analysis on majority training groups:}

On performing metadata mapping of the drivers for the largest consistent training group (group 1), we observe driver D4, the only female driver is the highest contributor i.e. has maximum instances in the group.  Table \ref{grp1} depicts the composition of group 1 by the driver information in number of windows. We observe for D4, almost all its normal (99\%) and drowsy (96\%) instances are present in this group (from Table 3 and 7). Similarly, normal and aggressive behavior of D5 and aggressive and drowsy behavior of D1 and D6 have dominant contribution to the group.

\begin{table}[htp]
\centering
\begin{tabular}{c| c c c |c}
\hline
\multirow{2}{*}{\textbf{Driver}}	& \multicolumn{3}{|c|}{\textbf{Driving Behavior (Group 1)}} & \multirow{2}{*}{\textbf{Total}}\\ \cline{2-4}
& Normal	& Agg &	Drowsy & \\ \hline
D1	& 98	& 168	& 195	& 461 \\
D2	& 116	& 104	& 125	& 345 \\
D3	& 115	& 35	& 132	& 282 \\
\textbf{D4}	& 252	& 188	& 249	& \textbf{689} \\
D5	& 224	& 188	& 9	& 421 \\
D6	& 140	& 165	& 255	& 550 \\
\hline
\end{tabular}
\caption{Composition of largest training group (group 1) by driver information and their driving behaviors}
\label{grp1}
\end{table}

Similarly, performing the same analysis on training group 2, depicted in table \ref{grp2}, we observe drivers D2 and D3 contributes maximum in forming this group. Together, they contribute to approximately 56\% of the entire group. Incidentally, from the metadata information, we observe both the drivers are of same age group (20-30) and have same fuel type of vehicle (Diesel) which may contribute to the similarity of their driving patterns.

\begin{table}[htp]
\centering
\begin{tabular}{c| c c c |c}
\hline
\multirow{2}{*}{\textbf{Driver}}	& \multicolumn{3}{|c|}{\textbf{Driving Behavior (Group 2)}} & \multirow{2}{*}{\textbf{Total}}\\ \cline{2-4}
& Normal	& Agg &	Drowsy & \\ \hline
D1	& 102	& 2	& 34	& 138 \\
\textbf{D2}	& 106	& 106	& 108	& \textbf{320} \\
\textbf{D3}	& 115	& 94	& 122	& \textbf{331} \\
D4	& 1	& 41	& 9	& 51 \\
D5	& 6	& 4	& 131	& 141 \\
D6	& 118	& 58	& 16	& 192 \\
\hline
\end{tabular}
\caption{Composition of training group 2 by driver information and their driving behaviors}
\label{grp2}
\end{table}

Figure \ref{dom} depicts the consistent training groups in the representation space and the dominant demographic traits and driving patterns characterized in each of the groups. 

In summary, we observe each of the groups has contributions from the three driving behaviors (else the training of the groups would be biased), but the individual driver’s behavior patterns are characterized uniquely in the consistent groups. This enables training driving behavior-pattern specific model which captures the diversity existing in individuals based on either their demographic features or their pattern of driving.

\section{Conclusion}

In this work, we propose a novel approach of driving behavior analysis, exploiting representation learning to encode the compact information of multivariate IMU timeseries and subsequently, forming consistent groups to capture subtle variations of driving patterns of different drivers. We have applied consistent grouping method using the learned representation, exploiting the choice of best clustering along with a selection of best distance measure.  Consistent groups are used to train different instances of a classifier to learn driving pattern specific model. Similarly, test data is divided into consistent groups and the recommended best distance measure is used to map the best train-test groups. In case of UAH driveset data, Chebyshev distance is selected as the best distance measure and is used for mapping the test groups to the optimal training model for inferencing. Here optimal indicates most suited train-test consistent group pairs.  We further perform in-depth analysis and validate that, the consistent groups on learned representation, indeed captures the driving behavior patterns. Proposed method outperforms benchmark classification methods like MLSTM-FCN and Stacked-LSTM which are trained using the complete training data. 	
In future, we plan to pursue further analysis of driving patterns causally related with various in-body and external factors. 

\bibliography{aaai22}

\end{document}